\documentclass[11pt,a4paper]{article}
\usepackage[hyperref]{acl2017}
\usepackage{times}
\usepackage{latexsym}
\usepackage{tabularx}
\usepackage{graphicx}
\usepackage{booktabs}
\usepackage[utf8]{inputenc}

\usepackage{url}

\aclfinalcopy 

\title{Neural Question Answering at BioASQ 5B}

\author{Georg Wiese\textsuperscript{1,2}, Dirk Weissenborn\textsuperscript{2} and Mariana Neves\textsuperscript{1} \\
\textsuperscript{1} Hasso Plattner Institute, August Bebel Strasse 88, Potsdam 14482 Germany \\
\textsuperscript{2} Language Technology Lab, DFKI, Alt-Moabit 91c, Berlin, Germany \\
  {\tt georg.wiese@student.hpi.de,}  \\ {\tt  dirk.weissenborn@dfki.de, mariana.neves@hpi.de } 
}

\date{}

\begin{document}
\maketitle
\begin{abstract}
  This paper describes our submission to the 2017 BioASQ challenge.
  We participated in Task B, Phase B which is concerned with biomedical question answering (QA).
  We focus on factoid and list question, using an \emph{extractive} QA model, that is, we restrict our system to output substrings of the provided text snippets.
  At the core of our system, we use FastQA, a state-of-the-art neural QA system.
  We extended it with biomedical word embeddings and changed its answer layer to be able to answer list questions in addition to factoid questions.
  We pre-trained the model on a large-scale open-domain QA dataset, SQuAD, and then fine-tuned the parameters on the BioASQ training set.
  With our approach, we achieve state-of-the-art results on factoid questions and competitive results on list questions.
\end{abstract}

\section{Introduction}

BioASQ is a semantic indexing, question answering (QA) and information extraction challenge \cite{tsatsaronis2015overview}.
We participated in Task B of the challenge which is concerned with biomedical QA.
More specifically, our system participated in Task B, Phase B:
Given a \emph{question} and gold-standard \emph{snippets} (i.e., pieces of text that contain the answer(s) to the question), the system is asked to return a list of answer candidates.

The fifth BioASQ challenge is taking place at the time of writing.
Five batches of $100$ questions each were released every two weeks.
Participating systems have 24 hours to submit their results.
At the time of writing, all batches had been released.

The questions are categorized into different question types: factoid, list, summary and yes/no.
Our work concentrates on answering \emph{factoid} and \emph{list} questions.
For factoid questions, the system's responses are interpreted as a ranked list of answer candidates.
They are evaluated using mean-reciprocal rank (MRR).
For list questions, the system's responses are interpreted as a set of answers to the list question.
Precision and recall are computed by comparing the given answers to the gold-standard answers.
F1 score, i.e., the harmonic mean of precision and recall, is used as the official evaluation measure \footnote{The details of the evaluation can be found at \url{http://participants-area.bioasq.org/Tasks/b/eval_meas/}}.

Most existing biomedical QA systems employ a traditional QA pipeline, similar in structure to the baseline system by \newcite{weissenborn2013answering}.
They consist of several discrete steps, e.g., named-entity recognition, question classification, and candidate answer scoring.
These systems require a large amount of resources and feature engineering that is specific to the biomedical domain.
For example, OAQA \cite{oaqa}, which has been very successful in last year's challenge, uses a biomedical parser, entity tagger and a thesaurus to retrieve synonyms.

Our system, on the other hand, is based on a neural network QA architecture that is trained end-to-end on the target task.
We build upon FastQA \cite{weissenborn2017fastqa}, an extractive factoid QA system which achieves state-of-the-art results on QA benchmarks that provide large amounts of training data.
For example, SQuAD \cite{rajpurkar2016squad} provides a dataset of $\approx100,000$ questions on Wikipedia articles.
Our approach is to train FastQA (with some extensions) on the SQuAD dataset and then fine-tune the model parameters on the BioASQ training set.

Note that by using an extractive QA network as our central component, we restrict our system's responses to substrings in the provided snippets.
This also implies that the network will not be able to answer yes/no questions.
We do, however, generalize the FastQA output layer in order to be able to answer list questions in addition to factoid questions.

\section{Model}

\begin{figure}[t]
\label{fig:biomedical_qa}
\centering
\includegraphics[width=0.5\textwidth]{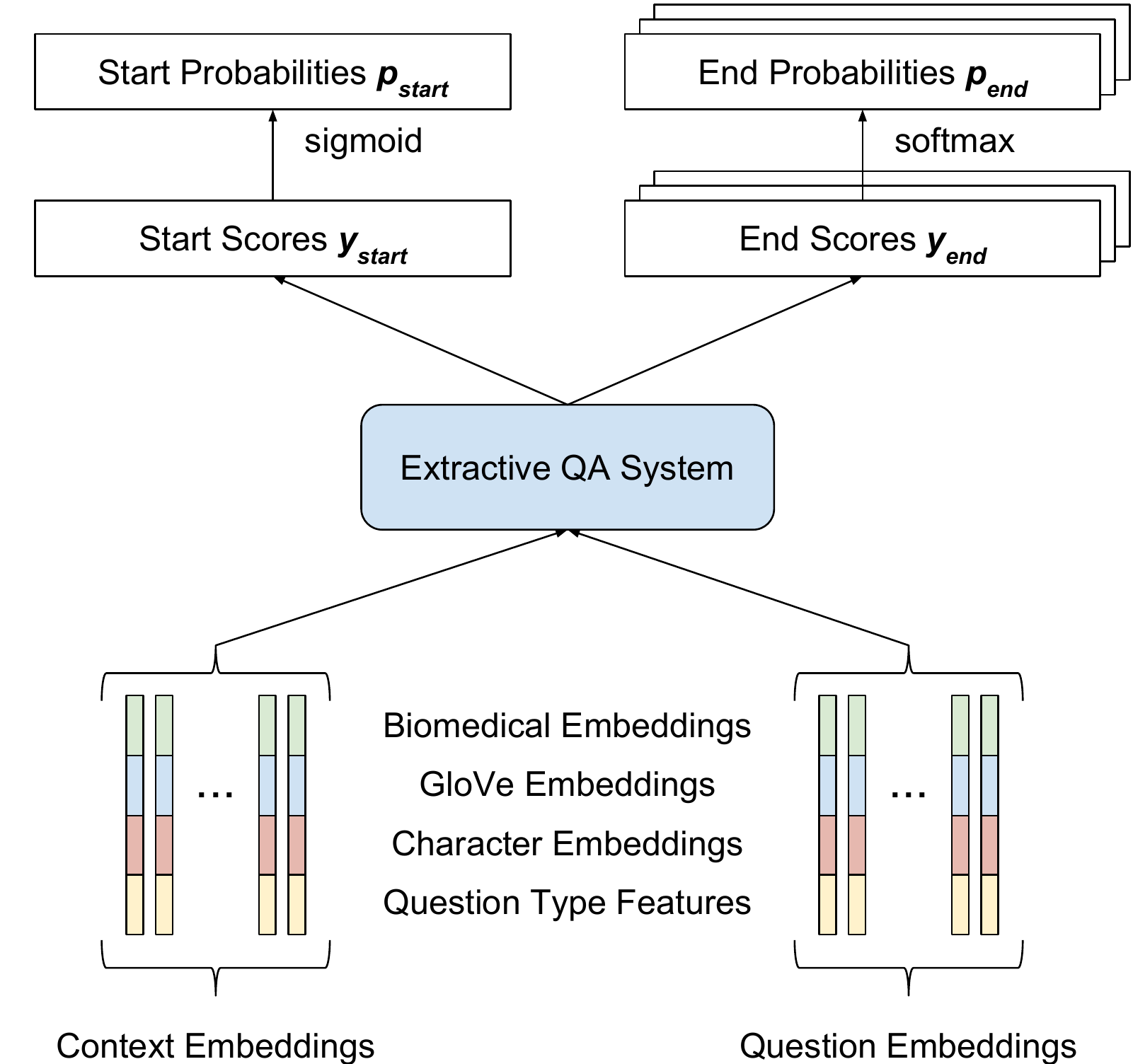}
\caption{Neural architecture of our system. Question and context (i.e., the \emph{snippets}) are mapped directly to start and end probabilities for each context token. We use FastQA \protect\cite{weissenborn2017fastqa} with modified input vectors and an output layer that supports list answers in addition to factoid answers.}
\end{figure}

Our system is a neural network which takes as input a question and a context (i.e., the snippets) and outputs start and end pointers to tokens in the context.
At its core, we use FastQA \cite{weissenborn2017fastqa}, a state-of-the-art neural QA system.
In the following, we describe our changes to the architecture and how the network is trained.

\subsection{Network architecture}

In the input layer, the context and question tokens are mapped to high-dimensional word vectors.
Our word vectors consists of three components, which are concatenated to form a single vector:

\begin{itemize}
\item
\textbf{GloVe embedding:} We use 300-dimensional GloVe embeddings \footnote{We use the \texttt{840B} embeddings available here: \url{https://nlp.stanford.edu/projects/glove/}} \cite{pennington2014glove} which have been trained on a large collection of web documents.
\item
\textbf{Character embedding:}
This embedding is computed by a 1-dimensional convolutional neural network from the characters of the words, as introduced by \newcite{seo2016bidirectional}.
\item
\textbf{Biomedical Word2Vec embeddings:}
We use the biomedical word embeddings provided by \newcite{biomedical_word2vec}.
These are 200-dimensional Word2Vec embeddings \cite{mikolov2013distributed_word2vec} which were trained on $\approx10$ million PubMed abstracts. 
\end{itemize}

To the embedding vectors, we concatenate a one-hot encoding of the question type (\emph{list} or \emph{factoid}).
Note that these features are identical for all tokens.

Following our embedding layer, we invoke FastQA in order to compute start and end scores for all context tokens.
Because end scores are conditioned on the chosen start, there are $O(n^2)$ end scores where $n$ is the number of context tokens.
We denote the start index by $i \in [1, n]$, the end index by $j \in [i, n]$, the start scores by $y_{start}^{i}$, and end scores by $y_{end}^{i, j}$.

In our output layer, the start, end, and span probabilities are computed as:

\begin{equation}
\label{eq:p_s}
	p_{start}^i = \sigma(y_{start}^i)
\end{equation}

\begin{equation}
\label{eq:p_e}
	p_{end}^{i, \cdot} = softmax(y_{end}^{i, \cdot})
\end{equation}

\begin{equation}
\label{eq:p_span}
	p_{span}^{i, j} = p_{start}^i \cdot p_{end}^{i, j}
\end{equation}

\noindent
where $\sigma$ denotes the sigmoid function.
By computing the start probability via the sigmoid rather than softmax function (as used in FastQA), we enable the model to output multiple spans as likely answer spans.
This generalizes the factoid QA network to list questions.

\begin{table*}[t]
  \centering
\begin{tabular}{l l l l l}
\toprule
\multicolumn{1}{c}{} & \multicolumn{2}{c}{Factoid MRR} & \multicolumn{2}{c}{List F1} \\
  \textbf{Batch} &  \textbf{Single} & \textbf{Ensemble} & \textbf{Single} & \textbf{Ensemble} \\
\midrule
1 & $52.0\%$ (2/10) & $57.1\%$ (1/10) & $33.6\%$ (1/11) & $33.5\% (2/11)$ \\
2 & $38.3\%$ (3/15) & $42.6\%$ (2/15) & $29.0\%$ (8/15) & $26.2\% (9/15)$ \\
3 & $43.1\%$ (1/16) & $42.1\%$ (2/16) & $41.5\%$ (2/17) & $49.5\% (1/17)$ \\
4 & $30.0\%$ (3/20) & $36.1\%$ (1/20) & $24.2\%$ (5/20) & $29.3\% (4/20)$ \\
5 & $39.2\%$ (3/17) & $35.1\%$ (4/17) & $36.1\%$ (4/20) & $39.1\% (2/20)$ \\
\hline
Average & $40.5\%$ & $42.6\%$ & $32.9\%$ & $35.1\%$ \\
\bottomrule
\end{tabular}
  \caption{
  Preliminary results for factoid and list questions for all five batches and for our single and ensemble systems.
  We report MRR and F1 scores for factoid and list questions, respectively.
  In parentheses, we report the rank of the respective systems relative to all other systems in the challenge.
  The last row averages the performance numbers of the respective system and question type across the five batches.
  }
  \label{tab:results}
\end{table*}

\subsection{Training \& decoding}
\label{sec:optimization_da}

\paragraph{Loss}
We define our loss as the cross-entropy of the correct start and end indices.
In the case of multiple occurrences of the same answer, we only minimize the span of the lowest loss.

\paragraph{Optimization}
We train the network in two steps:
First, the network is trained on SQuAD, following the procedure by \newcite{weissenborn2017fastqa} (\emph{pre-training phase}).
Second, we fine-tune the network parameters on BioASQ (\emph{fine-tuning phase}).
For both phases, we use the Adam optimizer \cite{kingma2014adam} with an exponentially decaying learning rate.
We start with learning rates of $10^{-3}$ and $10^{-4}$ for the pre-training and fine-tuning phases, respectively. 

\paragraph{BioASQ dataset preparation}
During fine-tuning, we extract answer spans from the BioASQ training data by looking for occurrences of the gold standard answer in the provided snippets.
Note that this approach is not perfect as it can produce false positives (e.g., the answer is mentioned in a sentence which does not answer the question) and false negatives (e.g., a sentence answers the question, but the exact string used is not in the synonym list).

Because BioASQ usually contains multiple snippets for a given question, we process all snippets independently and then aggregate the answer spans, sorting globally according to their probability $p_{span}^{i, j}$.

\paragraph{Decoding}
During the inference phase, we retrieve the top $20$ answers span via beam search with beam size $20$.
From this sorted list of answer strings, we  remove all duplicate strings.
For factoid questions, we output the top five answer strings as our ranked list of answer candidates.
For list questions, we use a \emph{probability cutoff threshold} $t$, such that $\{(i, j)|p_{span}^{i, j} \ge t\}$ is the set of answers.
We set $t$ to be the threshold for which the list F1 score on the development set is optimized.

\paragraph{Ensemble}
In order to further tweak the performance of our systems, we built a model ensemble.
For this, we trained five single models using 5-fold cross-validation on the entire training set.
These models are combined by averaging their start and end scores before computing the span probabilities (Equations~\ref{eq:p_s}-\ref{eq:p_span}).
As a result, we submit two systems to the challenge: The best single model (according to its development set) and the model ensemble.

\paragraph{Implementation}
We implemented our system using TensorFlow \cite{abadi2016tensorflow}.
It was trained on an NVidia GForce Titan X GPU.

\section{Results \& discussion}

We report the results for all five test batches of BioASQ 5 (Task 5b, Phase B) in Table~\ref{tab:results}.
Note that the performance numbers are not final, as the provided synonyms in the gold-standard answers will be updated as a manual step, in order to reflect valid responses by the participating systems.
This has not been done by the time of writing\footnote{The final results will be published at \url{http://participants-area.bioasq.org/results/5b/phaseB/}}.
Note also that -- in contrast to previous BioASQ challenges -- systems are no longer allowed to provide an own list of synonyms in this year's challenge.

In general, the single and ensemble system are performing very similar relative to the rest of field:
Their ranks are almost always right next to each other.
Between the two, the ensemble model performed slightly better on average.

On factoid questions, our system has been very successful, winning three out of five batches.
On list questions, however, the relative performance varies significantly.
We expect our system to perform better on factoid questions than list questions, because our pre-training dataset (SQuAD) does not contain any list questions.

Starting with batch $3$, we also submitted responses to yes/no questions by always answering \emph{yes}.
Because of a very skewed class distribution in the BioASQ dataset, this is a strong baseline.
Because this is done merely to have baseline performance for this question type and because of the naivety of the method, we do not list or discuss the results here.

\section{Conclusion}

In this paper, we summarized the system design of our BioASQ 5B submission for factoid and list questions.
We use a neural architecture which is trained end-to-end on the QA task.
This approach has not been applied to BioASQ questions in previous challenges.
Our results show that our approach achieves state-of-the art results on factoid questions and competitive results on list questions.

\bibliography{emnlp2016}
\bibliographystyle{acl_natbib}

\end{document}